\documentclass{article}
\usepackage{color,soul}
\usepackage{array}
\usepackage{times}
\usepackage{epsfig}
\usepackage{caption}
\usepackage[T1]{fontenc}
\usepackage[utf8]{inputenc}
\usepackage[english]{babel}
\usepackage[labelformat=simple]{subfig}
 
\usepackage{amsmath}
\usepackage{amssymb}
\usepackage{blindtext, subfig}
\usepackage{mathtools}
\usepackage{gensymb}
\usepackage{array}
\usepackage{times}
\usepackage{microtype}
\usepackage{graphicx}
\usepackage{booktabs} 
\usepackage[breaklinks=true,bookmarks=false]{hyperref}
 \usepackage{multirow}

\usepackage{hyperref}

\usepackage[accepted]{icml2019}

\icmltitlerunning{Unsupervised Temperature Scaling: An Unsupervised Post-Processing Calibration Method of Deep Networks}

\begin{document}

\twocolumn[
\icmltitle{Unsupervised Temperature Scaling:\\  An Unsupervised Post-Processing Calibration Method of Deep Networks}




\begin{icmlauthorlist}
\icmlauthor{Azadeh Sadat Mozafari}{to}
\icmlauthor{Hugo Siqueira Gomes}{to}
\icmlauthor{Wilson Leão}{goo}
\icmlauthor{Christian Gagn\'e}{to}
\end{icmlauthorlist}

\icmlaffiliation{to}{Computer Vision and Systems Laboratory, Universit\'e Laval, Quebec City, Canada}
\icmlaffiliation{goo}{Petrobras, Brazil}

\icmlcorrespondingauthor{Azadeh Sadat Mozafari}{azadeh-sadat.mozafari.1@ulaval.ca}

\icmlkeywords{Deep Neural Networks, Calibration, Temperature Scaling}

\vskip 0.3in
]



\printAffiliationsAndNotice{}

\begin{abstract}
The great performances of deep learning are undeniable, with impressive results over a wide range of tasks. However, the output confidence of these models is usually not well-calibrated, which can be an issue for applications where confidence on the decisions is central to providing trust and reliability (e.g., autonomous driving or medical diagnosis). For models using softmax at the last layer, Temperature Scaling (TS) is a state-of-the-art calibration method, with low time and memory complexity as well as demonstrated effectiveness. TS relies on a $T$ parameter to rescale and calibrate values of the softmax layer, whose parameter value is computed from a labelled dataset. We are proposing an Unsupervised Temperature Scaling (UTS) approach, which does not depend on labelled samples to calibrate the model, which allows, for example, the use of a part of a test samples to calibrate the pre-trained model before going into inference mode. We provide theoretical justifications for UTS and assess its effectiveness on a wide range of deep models and datasets. We also demonstrate calibration results of UTS on skin lesion detection, a problem where a well-calibrated output can play an important role for accurate decision-making.
\end{abstract}

\section{Introduction}
\label{Introduction}
Deep Neural Networks (DNNs) have demonstrated dramatically accurate results for challenging tasks \cite{he2016deep,simonyan2014very,graves2013speech,bar2015chest}. However, in real-world decision-making applications, accuracy is not the only element considered, the confidence of the network being also essential for a secure and reliable system. In DNNs, confidence usually corresponds to the output of a softmax layer, which is typically interpreted as the likeliness (probability) of different class occurrences. Most of the time, this value is far from the true probability of each class occurrence, with a tendency to become overconfident (i.e., output of one class being close to 1 the other outputs being close to 0). In such a case, we usually consider that the DNN not to be well-calibrated.

Calibration in DNNs is a challenge in deep learning, although it was not such an acute issue for shallow neural networks \cite{niculescu2005predicting}. \citet{guo2017calibration} has studied the role of different parameters leading to uncalibrated neural networks. They show that a deep network optimized with the Negative Log Likelihood (NLL) loss can reach higher accuracy when overfitting to NLL. However, the side effect of this is to make the network overconfident. 

Calibration plays an important role in real-world applications. In a self-driving car system \cite{bojarski2016end}, deciding weather to transfer the control of the car to the human observer relies on classification confidence over the detected objects. In medical care systems \cite{jiang2011calibrating}, false negative over deadly diseases can be catastrophic and as such, a proper confidence level is required for accurate decision-making. Calibration adds more information for an increased decision-making reliability.

Calibration methods for DNNs are widely investigated in recent literature and can be categorized into two main categories: 1) probabilistic approach; and 2) measurement-based approach. Probabilistic approaches generally include approximated Bayesian formalism \cite{ mackay1992bayesian,neal2012bayesian,louizos2016structured,gal2016dropout,lakshminarayanan2017simple}. In practice, the quality of predictive uncertainty in Bayesian-based methods relies heavily on the accuracy of sampling approximation and correctly estimated prior distribution. They are also suffering from time and memory complexity \cite{khan2018fast}.

Comparatively, measurement-based approaches are easier to use. They are generally post-processing methods that do not need to retrain the network in oreder to calibrate it . They apply a function on the logit layer of the network and fine-tune the parameters of that function by minimizing a calibration measure as a loss function. \citet{guo2017calibration} have compared several well-known measurement-based calibration methods such as Temperature, Matrix, Vector Scaling \cite{platt1999probabilistic}, Histogram Binning \cite{zadrozny2001obtaining}, Isotonic Regression \cite{zadrozny2002transforming} and Bayesian Binning into Quantiles \cite{naeini2015obtaining}.

Currently, Temperature Scaling (TS) \cite{guo2017calibration} is the state-of-the-art measurement-based approach when compared to the others. It achieves better calibration with minimum computational complexity (optimizing only one parameter $T$ to soften the softmax) compared to many approaches. It also preserves the accuracy rate of the network by not modifying the ranking of the network outputs. In TS, the best $T$ parameter is found by minimizing the Negative Log Likelihood (NLL) loss according to $T$ on a small-size labelled validation set. To calibrate a pre-trained model, gathering a labelled validation set is a costly procedure. Therefore, the possibility of calibrating the model with a small portion of the unlabelled samples from the test set can be an interesting post-processing solution.     

\textbf{Contributions}\quad In this paper, we propose a new TS family method which is called Unsupervised Temperature Scaling (UTS) to calibrate the network without accessing labels of the samples. Compared to TS algorithm, UTS preserves the minimal time and memory complexity advantage of classic TS as well as intact accuracy while allowing calibration during the inference phase using unlabelled samples. 

\section{Problem Setting}
\label{Problem Setting}

In this section, we define the problem, notations and introduce NLL~\cite{friedman2001elements} as a calibration measures.

\textbf{Assumptions}\quad We assume, we have access to a pre-trained deep model $D(\cdot)$ with the ability of detecting $K$ different classes. $D(\cdot)$ is trained on samples generated from distribution function $Q(\mathbf{x},y)$. We also have access to test data with the same distribution; we select $N$ number of samples from this test set as the validation set $\mathcal{V}= \{(\mathbf{x}_i)\}_{i=1}^N$ for calibration. 
For each sample $\mathbf{x}_i$, there exist $\mathbf{h}_i=[h_i^1,h_i^2,\ldots,h_i^K]^\top$ which is the logit layer. $D(\mathbf{x}_i) = (\hat{y_i},S_{y = \hat{y_i}}(\mathbf{x}_i))$ defines that the network $D(\cdot)$ detects label $\hat{y_i}$ for input sample $\mathbf{x}_i$ and confidence $S_{y=\hat{y_i}}(\mathbf{x}_i)$.  $S_y(\mathbf{x})=\exp({h_i^{{y}}})/ \sum_{j=1}^K \exp({h_i^j})$ is the softmax output function of the model that here is interpreted as the confidence.

\textbf{Goal}\quad The objective is to re-scale $\mathbf{h}_i$ according to parameter $T$ in order to minimize the calibration error of the model. 

\subsection{Negative Log Likelihood (NLL)}
When the network is well-calibrated, the softmax output layer should indicate the true conditional distribution $Q(y|\mathbf{x})$. Therefore, to measure calibration, we are looking at the similarity between the softmax output $S_y(\mathbf{x})$ and true distribution function $Q(y|\mathbf{x})$. As the true distribution function $Q(y|\mathbf{x})$ is not available, Gibbs inequality can be helpful to find the similarity between two distributions. Gibbs inequality is valid for any arbitrary distribution function $P(y|\mathbf{x})$ as:
\begin{equation}
-\mathbb{E}_{Q(\mathbf{x},y)}[\log \left( Q(y|\mathbf{x})\right)] \leq -\mathbb{E}_{Q(\mathbf{x},y)}[\log \left(P(y|\mathbf{x})\right)],
\label{eq1}
\end{equation}
where $\mathbb{E}$ is the expected value function on distribution $Q(\mathbf{x},y)$. 
The minimum of $-\mathbb{E}_{Q(\mathbf{x},y)}[\log P(y|\mathbf{x})]$ occurs when $P(y|\mathbf{x})$ is equal to $Q(y|\mathbf{x})$. NLL is defined as the empirical estimation of $-\mathbb{E}_{Q(\mathbf{x},y)}[\log P(y|\mathbf{x})]$ which can be rephrased as distribution function $S_{y}(\mathbf{x})$ of a deep neural network:
\begin{equation}
\text{NLL} = -\sum_{(\mathbf{x}_i,y_i)}\log \left(S_{y=y_i}(\mathbf{x}_i)\right), \quad  (\mathbf{x}_i,y_i)\sim Q(\mathbf{x},y).
\label{eq2}
\end{equation}
To minimize the calibration error, NLL will be minimized with respect to the parameters that fine-tune the
$S_y(\mathbf{x})$ function. This is equal to minimize the distance between DNN output confidence and the true distribution $Q(y|\mathbf{x})$.

\section{Temperature Scaling (TS)}
\label{Temperature Scaling}

TS has previously been applied for calibration \cite{guo2017calibration}, distilling knowledge \cite{hinton2015distilling} and enhancing the output of DNNs for better discrimination between the in- and out-of-distribution samples \cite{liang2017enhancing}. TS is a post-processing approach that rescales the logit layer of a deep model through temperature $T$. TS is used to soften the output of the softmax layer and  makes it better calibrated. The best value of $T$ will be obtained by minimizing NLL loss function according to $T$ on a small labelled dataset  $\mathcal{T}=\{(\mathbf{x}_i,y_i)\sim Q(\mathbf{x},y)\}_{i=1}^N$:
\begin{align}
T^* =&\operatorname*{arg\,min}_{T} \left(-\sum_{i=1}^N\log\big(S_{y = y_i}(\mathbf{x}_i,T)\big)\right)\label{Eq(5)}\\    
s.t: & \quad T>0,  \hspace {3mm} (\mathbf{x}_i,y_i)\in \mathcal{T},\nonumber
\end{align}
where $S_{y=y_i}(\mathbf{x}_i,T)= \exp({\frac{h_i^{y_i}}{T}})/ \sum_{j=1}^K \exp(\frac{h_i^j}{T})$, is a rescaled version of softmax by applying temperature $T$. TS has the minimum time and memory complexity among calibration approaches as it only requires the optimization of one parameter $T$ on a small validation set. Having only one parameter not only helps TS to be efficient and practical but also to avoid overfitting to NLL loss function when it is optimized on a small validation set $\mathcal{T}$. TS also preserves the accuracy intact during the calibration which can be an important feature when there is a risk of an accuracy drop because of calibration. 

\section{Unsupervised Temperature Scaling (UTS)}
\label{Unsupervised Temperature Scaling}

\begin{figure*}[tb]
\centering

\includegraphics[height = 6cm]{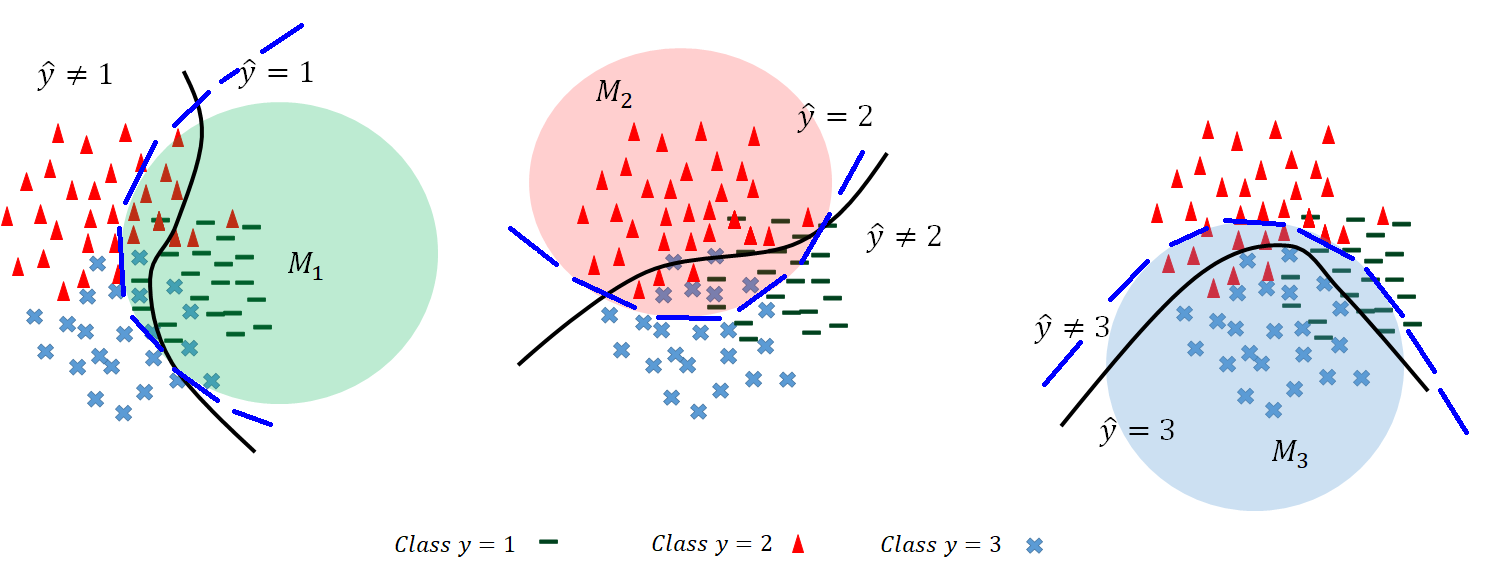}

\caption{$M_k$ subset division for a three-class classification problem. The samples generated from distribution $Q(\mathbf{x},y=k)$ are located in colored region of $M_k$, where $S_{y=k}(\mathbf{x}_i)\geq\theta_k$. The dashed blue line shows the boundary of $S_{y=k}(\mathbf{x})=\theta_k$ and the black continuous line shows the decision boundary which discriminates samples based on the label prediction $\hat{y}$.} 

\label{figure_M_division}
\end{figure*}

This section describes our proposed approach which we refer to as Unsupervised TS. It minimizes NLL according to the $T$ parameter for each class distribution instead of the total data distribution. This class-oriented distribution matching helps UTS to get rid of the true label of the samples. More specifically, to estimate each class distribution, UTS minimizes NLL loss of $S_{y=k}(\mathbf{x})$ with respect to $T$ on the samples assumed to be generated from $Q(\mathbf{x},y=k)$. The UTS loss function is proposed as Eq.~\ref{Eq(8)} and the best temperature will be obtained from Eq.~\ref{Eq(temperature)}:
\begin{align}
\mathcal{L}_{UTS} & = \sum_{k=1}^{K}\sum_{\mathbf{x}_i\in M_k} -\log\Big(S_{y=k}(\mathbf{x}_i,T)\Big),\label{Eq(8)}\\
T^* & = \operatorname*{arg\,min}_{T}(\mathcal{L}_{UTS}), \qquad s.t: T>0,\label{Eq(temperature)}
\end{align}
$M_k$ contains the samples which are highly probable to be generated from $Q(\mathbf{x},y=k)$ distribution. Generally the samples that have the true label $y=k$ should be in $M_k$. However UTS does not have access to the true labels of the samples. Accordingly, we are assuming that the output of the classifier can be used as an indicator to select the samples for $M_k$ (i.e., samples with the higher softmax value for $S_{y=k}(\mathbf{x})$ ). Formally $M_k$ is defined as:
\begin{equation}
M_k=\{\mathbf{x}_i ~ | ~ S_{y=k}(\mathbf{x}_i)\geq \theta_k , ~~  \mathbf{x}_i\in \mathcal{V}\},\label{Eq(7)}
\end{equation}
with $\theta_k$ being a threshold which is defined for each class $k$ to select the samples with higher $S_{y=k}(\mathbf{x})$ value. We will explain about the selecting threshold strategy in Sec.~\ref{threshold}. $M_k$ contains the samples that their true labels might be $y=k$ or $y\neq k$. Fig.~\ref{figure_M_division} shows the samples selected for $M_k$ in the case of three class classification problem. When the samples have the true label $k$, they are generated from $Q(\mathbf{x},y=k)$ which fulfills our final goal. In the case that their true label is $y\neq k$, we will show in Sec. \ref{sample_distribution}, it also can be considered as the sample generated from distribution $Q(x,y=k)$ if they are selected from the samples located near to the decision boundary of class $k$.

\begin{table*}[tb]
\centering
\resizebox{0.8\textwidth}{!}{  
\begin{tabular}{l|l|c|cc|ccc|ccc}
\toprule
                &           &           & \multicolumn{2}{c|}{\textbf{Uncalibrated}} & \multicolumn{3}{c|}{\textbf{TS}} & \multicolumn{3}{c}{\textbf{UTS}}\\
Model           & {Dataset} & Accuracy  & NLL   & ECE       & NLL               & ECE               & $T$               & NLL               & ECE               & $T$\\
\midrule
DenseNet40      & CIFAR10   & {92.61\%} & 0.286 & 4.089     & 0.234             & 3.241             & 2.505             & \textbf{0.221}    & \textbf{0.773}    & 1.899\\
DenseNet40      & CIFAR100  & {71.73\%} & 1.088 & 8.456     & \textbf{1.000}    & \textbf{1.148}    & 1.450             & {1.001}           & {1.945}           & 1.493\\
DenseNet100     & CIFAR10   & {95.06\%} & 0.199 & 2.618     & \textbf{0.156}    & \textbf{0.594}    & 1.801             & {0.171}           & {3.180}           & 2.489\\
DenseNet100     & CIFAR100  & {76.21\%} & 1.119 & 11.969    & 0.886             & 4.742             &2.178              & \textbf{0.878}    & \textbf{2.766}    & 1.694\\
DenseNet100     & SVHN      & {95.72}\% & 0.181 & 1.630     & {0.164}           & \textbf{0.615}    & 1.407             & \textbf{0.162}    & {1.074}           & 1.552\\ 
ResNet110       & CIFAR10   & {93.71\%} & 0.312 & 4.343     & 0.228             & 4.298             & 2.960             & \textbf{0.207}    & \textbf{1.465}    & 2.009\\
ResNet110       & CIFAR100  & {70.31\%} & 1.248 & 12.752    & \textbf{1.051}    & \textbf{1.804}    & 1.801             & {1.055}           & {2.796}           & 1.562\\
ResNet110       & SVHN      & {96.06\%} & 0.209 & 2.697     & 0.158             & 1.552             & 2.090             & \textbf{0.152}    & \textbf{0.550}    & 1.758\\
WideResNet32    & CIFAR100  & {75.41\%} & 1.166 & 13.406    & 0.909             & \textbf{4.096}    & 2.243             & \textbf{0.905}    & 4.872             & 1.651\\
LeNet5          & MNIST     & {99.03\%} & 0.105 & 0.727     & 0.061             & 0.674             & 1.645             & \textbf{0.043}    & \textbf{0.584}    & 1.857\\
VGG16           & CIFAR10   & {92.09\%} & 0.427 & 5.99      & 0.301             & 6.015             & 3.229             & \textbf{0.268}    & \textbf{1.675}    & 2.671\\
\bottomrule
\end{tabular}
}
\caption{The results of calibration for UTS vs. TS and uncalibrated models for variation of datasets and models. We report the results for two different calibration measures NLL and ECE (explanation on ECE is given in Appendix, Sec. \ref{ECE}). Smaller values indicate better calibration. We also report  $T$ values for TS and UTS to show that small changes in $T$ can lead to significant improvements in calibration.}
\vspace{-0.3cm}

\label{Tabel_calibration}
\end{table*}

\subsection{Distribution of samples near to the decision boundary}
\label{sample_distribution}

Let assume we have a classifier $k$ that discriminates the samples associated to class $k$ from samples not from that class. For the samples located near to the decision boundary of this classifier, the conditional distribution of them are defined as  $Q(y=k|\mathbf{x}) \simeq Q(y\not=k|\mathbf{x})$. From Bayes rule, we have:
\begin{equation}
Q(\mathbf{x},y=k) = \frac{Q(y=k|\mathbf{x})}{Q(y\neq k|\mathbf{x})}Q(\mathbf{x},y \neq k).\label{Eq(6)}
\end{equation}
Therefore, for the samples near to the decision boundary,as  they have $Q(y=k|\mathbf{x})/Q(y\neq k|\mathbf{x})\simeq1$, we can use them as the samples generated from distributions $Q(\mathbf{x},y=k)$ or $Q(\mathbf{x},y\not=k)$ interchangeably. 

\subsection{Selecting threshold $\theta_k$}
\label{threshold}
Threshold $\theta_k$ should be defined in order to select the samples that are more probable to be generated from $Q(\mathbf{x},y=k)$ which means the samples with higher confidence value for class $k$. We focus on the samples with $\hat{y} \neq k$ and define the threshold in order to select the upper range of ${S}_{y=k}(\mathbf{x})$ confidence for them. This threshold also will select the samples that are predicted as $\hat{y}=k$. Threshold $\theta_k$ is defined as below:
\begin{equation}
\theta_k = \bar{ S}_{y=k}+\sqrt{\frac{1}{|\mathcal{U}|}\sum_{\mathbf{x}_i\in\mathcal{U}}(\bar{S}_{y=k}-S_{y=k}(\mathbf{x}_i))^2}, \label{Eq(7)}
\end{equation}
  Eq.~\ref{Eq(7)} simply calculates the sum of the mean and standard deviation of ${S}_{y=k}(\mathbf{x})$ for the samples in  $\mathcal{U}$, where  $\mathcal{U} = \{\mathbf{x}_i|\hat{y}_i\not=k\}$. However how to find $\theta_k$ more precisely can be considered as the future work. 

\section{Experimental Results}

\begin{figure}[tb]
 \centering
 \subfloat[][\scriptsize Label =  Melanoma\\Pred. =  Melanoma\\Confidence = 0.99\\Calib. Confidence = 0.99 ]{\label{fig:1random}\includegraphics[height=2.4cm,width=2.7cm]{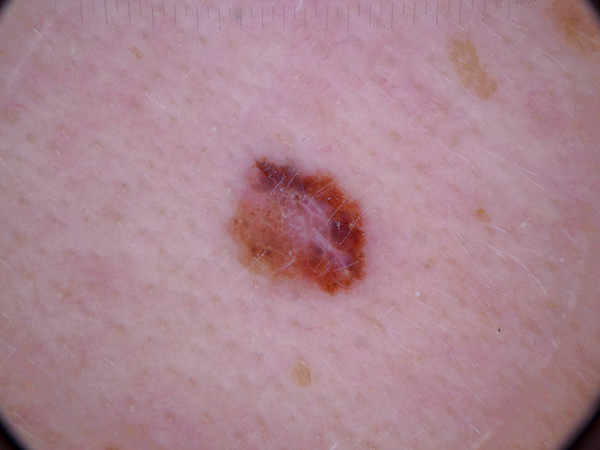}}
 \hfill
 \subfloat[][\scriptsize Label = Dermatofibroma\\Pred. =  Dermatofibroma\\Confidence = 0.99\\Calib. Confidence = 0.97 ]{\label{fig:2random}\includegraphics[height=2.4cm,width=2.7cm]{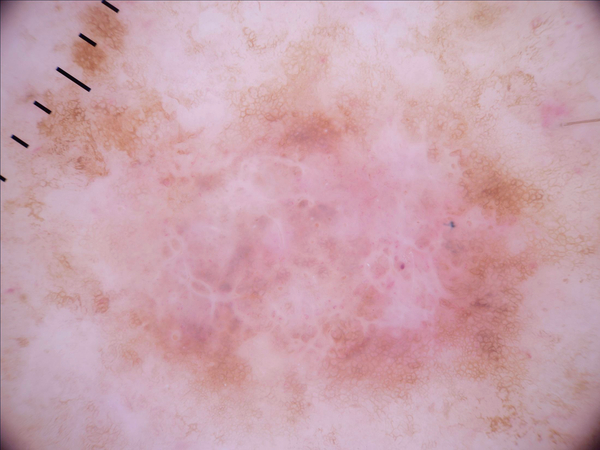}}
 \hfill
 \subfloat[][\scriptsize Label = BCC\\Pred. =  BCC\\Confidence = 0.99\\Calib. Confidence = 0.98 ]{\label{fig:3random}\includegraphics[height=2.4cm,width=2.7cm]{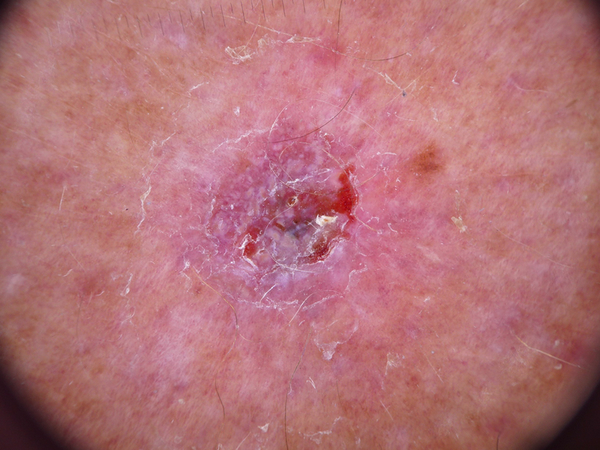}}\\
 
 \subfloat[][\scriptsize Label = Melanoma\\Pred. = Bowen\\Confidence = 0.88\\Calib. Confidence = 0.68 ]{\label{fig:4random}\includegraphics[height=2.4cm,width=2.7cm]{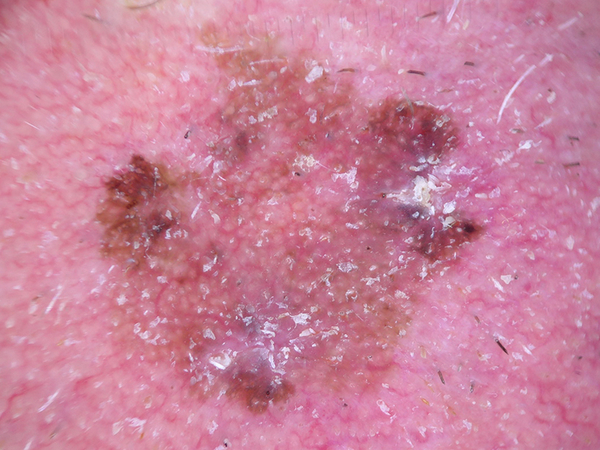}}
  \hfill
\subfloat[][\scriptsize Label = Bowen\\Pred. =  BCC\\Confidence = 0.90\\Calib. Confidence = 0.64 ]{\label{fig:5random}\includegraphics[height=2.4cm,width=2.7cm]{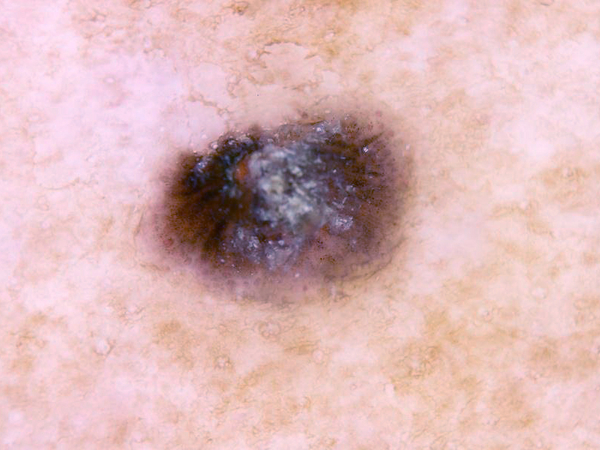}}
  \hfill
\subfloat[][\scriptsize Label = Melanocytic nevus\\Pred. =  Benign keratosis\\Confidence = 0.92\\Calib. Confidence = 0.68 ]{\label{fig:6random}\includegraphics[height=2.4cm,width=2.7cm]{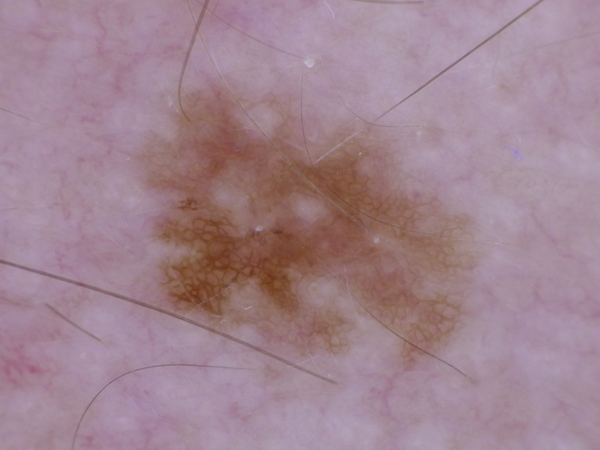}} 
\caption{Confidence of skin anomaly detection system before and after calibration with UTS. UTS decreases the confidence of misclassified samples while tends to preserve a high confidence over correctly classified samples. }
\vspace{-5mm}
\label{fig_calibration}
\end{figure}

We investigate the effectiveness of UTS in comparison to TS on a wide range of different model-dataset combinations. In all experiments, we divide randomly the test data into $20\%$ and $80\%$. We use $20\%$ portion of data to calibrate temperature and the reminder as test dataset to report the results on. For TS we use the labels to tune the temperature while for UTS we ignored them using samples as unlabeled instances. 

\textbf{Datasets}\quad To investigate the validity of UTS, we test the methods on CIFAR-10 \cite{krizhevsky2009learning}, CIFAR-100 \cite{krizhevsky2009learning}, SVHN \cite{netzer2011reading}, and MNIST \cite{lecun1998mnist}. For calibration experiments we use data extracted from the ``ISIC 2018: Skin Lesion Analysis Towards Melanoma Detection'' grand challenge datasets \cite{tschandl2018ham10000,DBLP:journals/corr/abs-1710-05006} that includes images of 7 types of skin lesions.

\textbf{Models}\quad We tested a wide range of different state-of-the-art deep convolutional networks with variations in depth. The selected DNNs are Resnet \cite{he2016deep}, WideResnet \cite{zagoruyko2016wide}, DenseNet \cite{iandola2014densenet}, LeNet \cite{lecun1998gradient}, and VGG \cite{simonyan2014very}. We use the data pre-processing, training procedures and hyper-parameters as described in each paper.

\subsection{Results for Different Model-Datasets}

Table \ref{Tabel_calibration} compares the calibration results of UTS to TS. In both methods the accuracy remains unchanged before and after calibration. UTS improves the calibration in several cases compared to TS and is always better than the corresponding uncalibrated model.  

\subsection{Calibration Application: Skin Lesion Detection}

UTS is also used to calibrate a real-world application where labeling is an expensive task. We trained ResNet152 on the ISIC dataset \cite{tschandl2018ham10000, DBLP:journals/corr/abs-1710-05006} to detect seven different skin lesions, with an accuracy of $88.02\%$ -- see Sec.~\ref{Implementation_details} and Sec.~\ref{More_results} of the Appendix for more details. Fig.~\ref{fig_calibration} shows the model confidences before and after calibration with UTS for correctly classified and misclassified samples. Confidence was high for both groups before calibration, while it was maintained high after calibration but much reduced for misclassified samples.


\bibliography{example_paper}
\bibliographystyle{icml2019}

\appendix

\section*{Appendix}

\section{Expected Calibration Error (ECE) }
\label{ECE}
Another way to define calibration is based on the relation between the accuracy and confidence. Miscalibration can be interpreted as the difference between confidence and probability of correctly classifying a sample. For instance, in the case of a calibrated model, if we have the group of samples which has the confidence of $S_{y}(\mathbf{x})=0.9$, it is supposed to have $0.9$ percentage of accuracy. Based on this definition of calibration, ECE is proposed as the empirical expectation error between the accuracy and confidence \cite{naeini2015obtaining} in a range of confidence intervals. 
It is calculated by partitioning the range of confidence which is between $[0\,,1]$ into $L$ equally-spaced confidence bins and then assign the samples to each bin $B_l$ where $l=\{1,\ldots,L\}$ by their confidence range. Later it calculates the weighted absolute difference between the accuracy and confidence for each subset $B_l$. More specifically:
\begin{equation}
\text{ECE} = \sum_{l=1}^L{\frac{|B_l|}{N}}\Big|\text{acc}(B_l)-\text{conf}(B_l)\Big|,
    \label{eq4}
\end{equation}\\
where $N$ is the total number of samples and $|B_l|$ is the number of samples falling into the interval $B_l$. 

\section{Implementation Specification of Skin Lesion Detection System}
\label{Implementation_details}
To test the impact of calibration in a real application, we design a medical assistant system. One of the medical applications is anomaly detection for skin spots. We use ISIC dataset \cite{tschandl2018ham10000,DBLP:journals/corr/abs-1710-05006} which contains 10015  images of 7 different skin lesion types  which are, Melanoma, Melanocytic nevus, Basal cell carcinoma (BCC), Bowen, Benign keratosis, Dermatofibroma, and  Vascular. We select  ResNet152 with pretrained weights on ImageNet as the classifier. In order to fine-tune it, we use 60\% of ISIC images resizing them to $224 \times 224$ and normalizing with mean and standard deviation of the ImageNet dataset. Notice that we use stratification to divide the dataset. We run the fine-tuning for 100 epochs with the batchsize of 32 using Adam optimizer and beginning with the learning rate of 1e-4 and a scheduled decaying rate of 0.95 every 10 epochs. To increase the variety of the training samples, we perform data augmentation with a probability of 0.5 of transforming every image with a random horizontal or vertical flip or a random rotation of a maximum of 12.5\degree ~either to the left or to the right.
From the $40\%$ of the data that we consider as the test samples, we select $20\%$ of them as the validation which is $801$ samples for calibration and hyper parameter tuning. We report the results on the rest of the test samples which are $3205$.  
\section{More Results of Skin Lesion Detection System}
\label{More_results}
\begin{figure*}[!ht]
 \centering
 \subfloat[][\scriptsize Label = Benign keratosis\\Pred. = Benign keratosis\\Confidence = 0.99\\UTS Confidence = 0.99 ]{\label{fig:gull}\includegraphics[height=3cm,width=3.2cm]{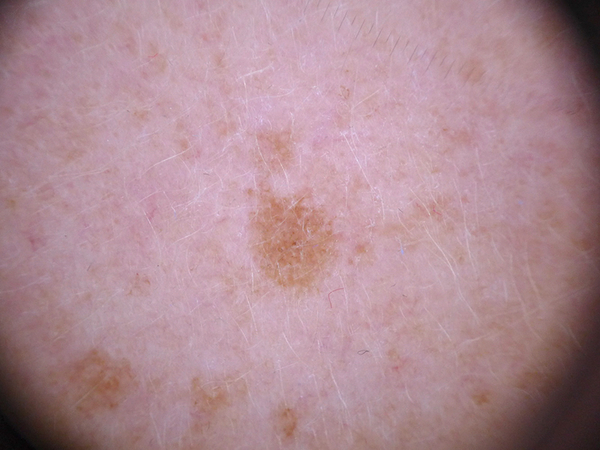}}\quad
 \subfloat[][\scriptsize Label = Melanocytic nevus\\Pred. =  Melanocytic nevus \\Confidence = 0.99\\UTS Confidence =  0.98 ]{\label{fig:gull}\includegraphics[height=3cm,width=3.2cm]{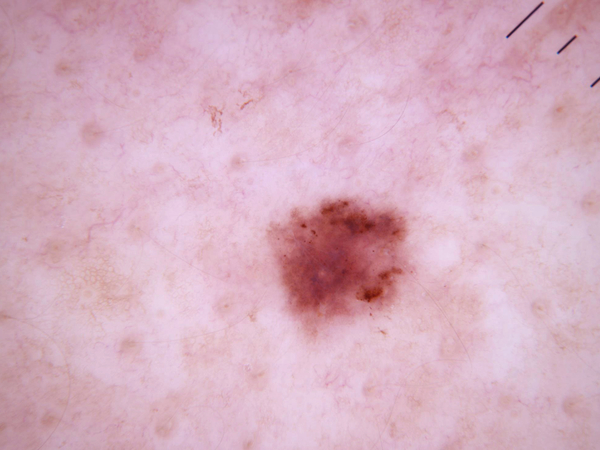}}\quad
 \subfloat[][\scriptsize Label = Bowen\\Pred. = Bowen\\Confidence =0.99 \\UTS Confidence = 0.98 ]{\label{fig:gull}\includegraphics[height=3cm,width=3.2cm]{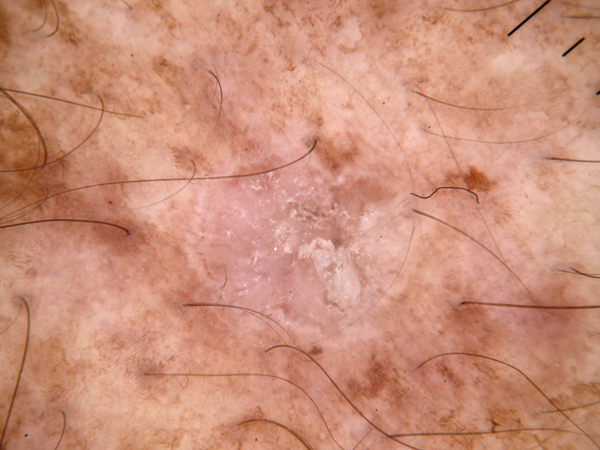}}\quad
 \subfloat[][\scriptsize Label = Dermatofibroma\\Pred. = Dermatofibroma \\Confidence =  0.99\\UTS Confidence = 0.97]{\label{fig:gull}\includegraphics[height=3cm,width=3.2cm]{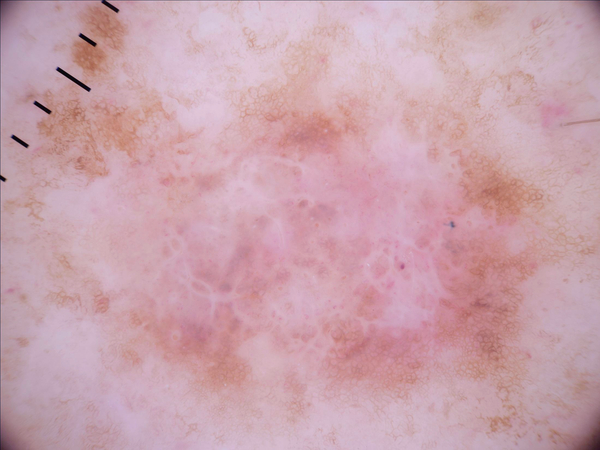}}\\
 \subfloat[][\scriptsize Label = BCC\\Pred. = BCC\\Confidence = 0.99\\UTS Confidence =  0.98 ]{\label{fig:gull}\includegraphics[height=3cm,width=3.2cm]{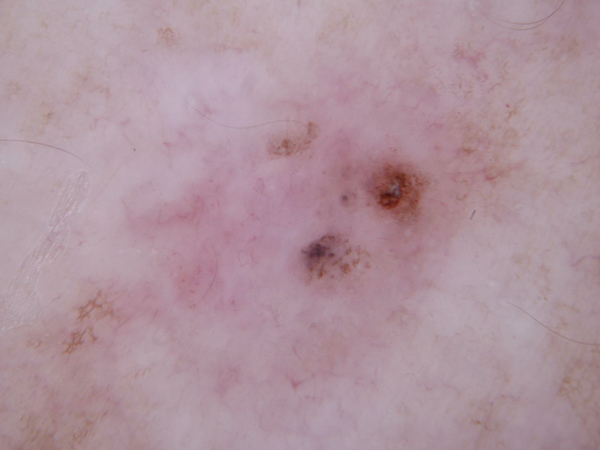}}\quad
  \subfloat[][\scriptsize Label = Melanocytic nevus\\Pred. = Melanocytic nevus\\Confidence = 0.99\\UTS Confidence = 0.98 ]{\label{fig:gull}\includegraphics[height=3cm,width=3.2cm]{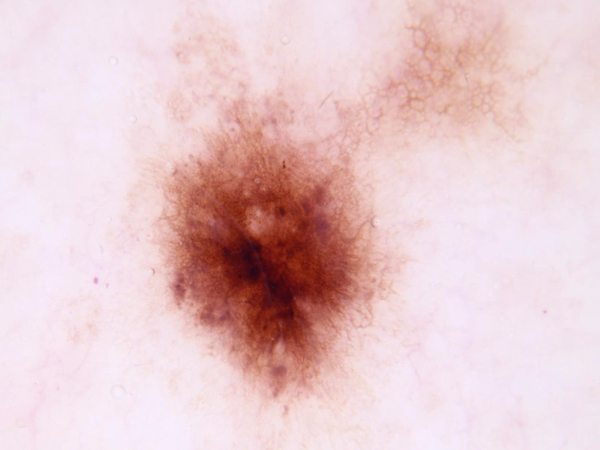}}\quad
 \subfloat[][\scriptsize Label = Vascular lesion\\Pred. = Vascular lesion\\Confidence = 0.99 \\UTS Confidence = 0.97 ]{\label{fig:gull}\includegraphics[height=3cm,width=3.2cm]{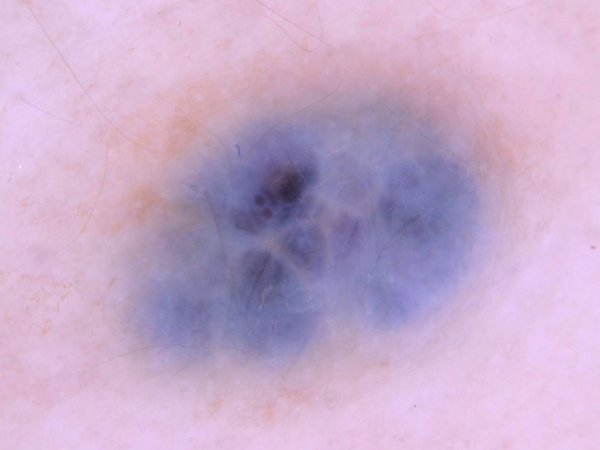}}\quad
 \subfloat[][\scriptsize Label =  Melanoma\\Pred. = Melanoma\\Confidence = 0.99\\UTS Confidence = 0.94 ]{\label{fig:gull}\includegraphics[height=3cm,width=3.2cm]{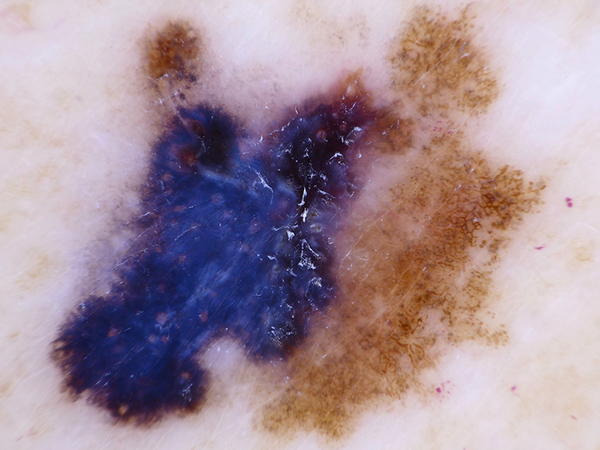}}\\
 \subfloat[][\scriptsize Label =  Melanoma\\Pred. = Melanocytic nevus\\Confidence = 0.90\\UTS Confidence = 0.78 ]{\label{fig:gull}\includegraphics[height=3cm,width=3.2cm]{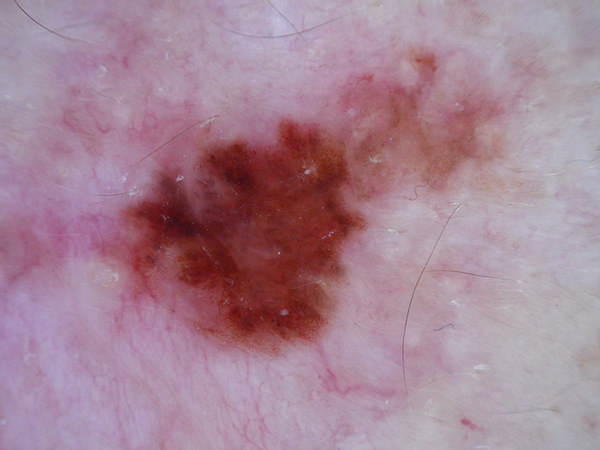}}\quad
 \subfloat[][\scriptsize Label = Benign keratosis\\Pred. = Melanocytic nevus\\Confidence = 0.90\\UTS Confidence = 0.77 ]{\label{fig:gull}\includegraphics[height=3cm,width=3.2cm]{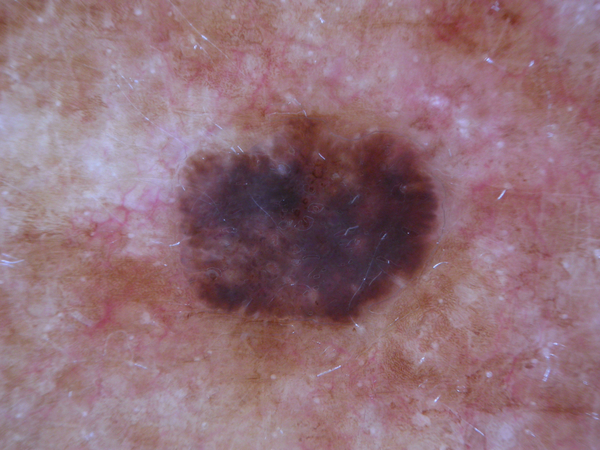}}\quad
  \subfloat[][\scriptsize Label = Melanocytic nevus\\Pred. = BCC\\Confidence = 0.90\\UTS Confidence = 0.79 ]{\label{fig:gull}\includegraphics[height=3cm,width=3.2cm]{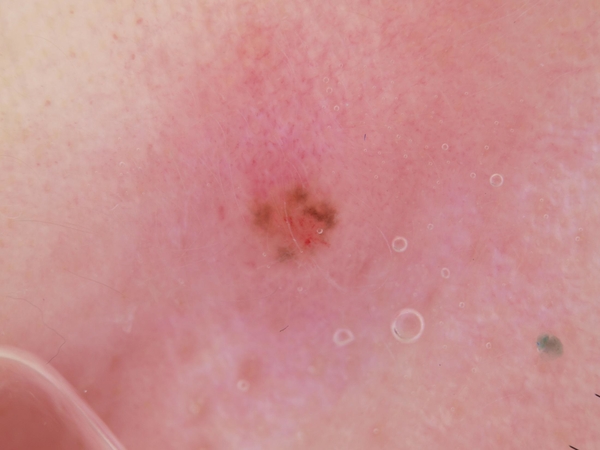}}\quad
 \subfloat[][\scriptsize Label = Melanocytic nevus\\Pred. = Benign keratosis\\Confidence = 0.90\\UTS Confidence =  0.78 ]{\label{fig:gull}\includegraphics[height=3cm,width=3.2cm]{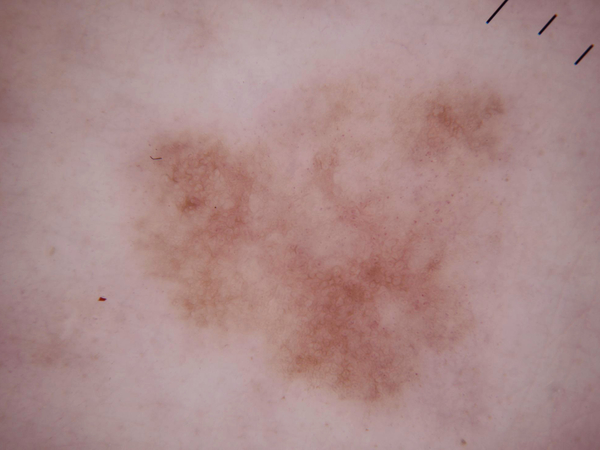}}\\
 \subfloat[][\scriptsize Label = Dermatofibroma\\Pred. = Benign keratosis\\Confidence = 0.95\\UTS Confidence = 0.79]{\label{fig:gull}\includegraphics[height=3cm,width=3.2cm]{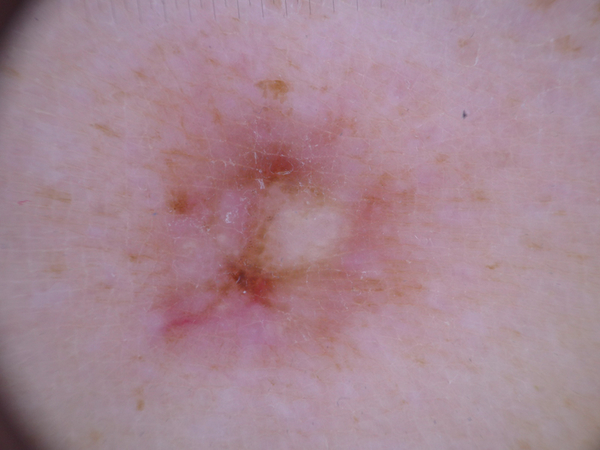}}\quad
  \subfloat[][\scriptsize Label = Bowen\\Pred. = Melanoma  \\Confidence = 0.91\\UTS Confidence = 0.76 ]{\label{fig:gull}\includegraphics[height=3cm,width=3.2cm]{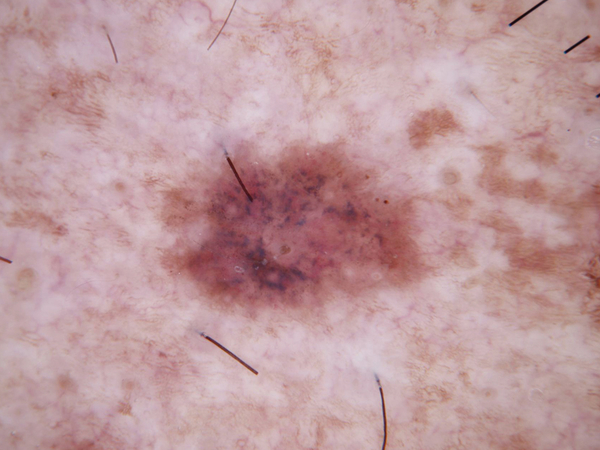}}\quad
 \subfloat[][\scriptsize Label = Melanoma\\Pred. = Bowen\\ Confidence = 0.92\\UTS Confidence = 0.79 ]{\label{fig:gull}\includegraphics[height=3cm,width=3.2cm]{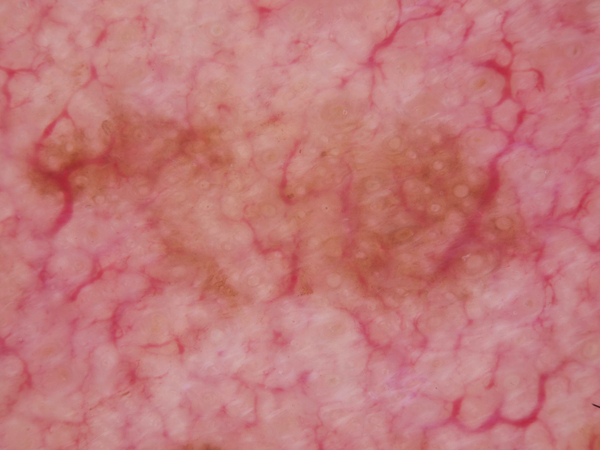}}\quad
 \subfloat[][\scriptsize Label = BCC\\Pred. = Bowen \\Confidence = 0.90 \\UTS Confidence =  0.78 ]{\label{fig:gull}\includegraphics[height=3cm,width=3.2cm]{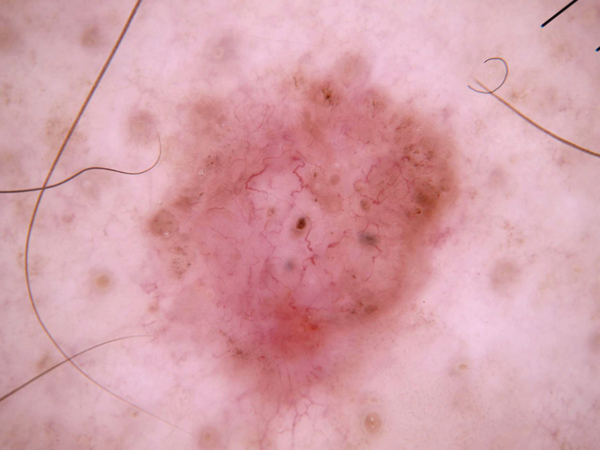}}\quad
\caption{Different correctly classified and misclassified output of skin lesion detection system before and after calibration with UTS. }
\label{Skin_Lesion_TS_UTS}
\end{figure*}

In this section, we provide more results of the skin lesion detection system. The confidence of the system before and after calibration by UTS method for correctly classified and misclassified samples is reported in Fig.~ \ref{Skin_Lesion_TS_UTS} for different skin lesion types. The temperature and calibration error of the trained model is reported in Table \ref{Tabel_calibration_ISIC}.

\begin{table*}[tb]
\centering

\begin{tabular}{c|c|c|c|c|c|c|c}
\toprule
            &           &           & \multicolumn{2}{c|}{\textbf{Uncalibrated}}   & \multicolumn{3}{c}{\textbf{UTS}}\\
Model       & Dataset   & Accuracy  & NLL   & ECE                           & NLL       & ECE   & $T$\\
\midrule
ResNet152   & ISIC      & 88.02\%   & 0.712 & 0.092                         &  0.461    & 0.068 & 1.712\\
\bottomrule
\end{tabular}
\caption{Comparing calibrating ResNet152 model trained on ISIC dataset with UTS approach. }
\label{Tabel_calibration_ISIC}
\end{table*}

\end{document}